%% file: wann.tex
\RequirePackage{fix-cm}

\documentclass[twocolumn]{svjour3}          
\smartqed  
\usepackage{graphicx}
\usepackage{optidef}

\usepackage{caption}
\usepackage{subcaption}

\usepackage{multirow,booktabs,color,soul}

\usepackage[ruled,linesnumbered]{algorithm2e}
\usepackage{amsmath}
\usepackage{color}
\usepackage{wrapfig}

\begin{document}

\title{Accelerating Multi-Objective Neural Architecture Search by Random-Weight Evaluation
}

\titlerunning{Accelerating Multi-Objective Neural Architecture Search by Random-Weight Evaluation}        

\author{
Shengran Hu \and
Ran Cheng$^*$ 
\and
Cheng He \and
Zhichao Lu \and
Jing Wang \and
Miao Zhang
}


\institute{
$^*$Ran Cheng is the corresponding author.\\ \\
S. Hu, R. Cheng, C. He, and Z. Lu \at
            Guangdong Provincial Key Laboratory of Brain-inspired Intelligent Computation, Department of Computer Science and Engineering, Southern University of Science and Technology, Shenzhen 518055, China.\\
            \email{
                hu.shengran@outlook.com, ranchengcn@gmail.com, 
                chenghehust@gmail.com, lu.zhichao@outlook.com
            }           
          \and
          J. Wang and M. Zhang \at
              Shanghai Aircraft Design and Research Institute, Shanghai 200135, China. \\
            \email{
              wangjinger1218@163.com, zhangm-168@163.com
            }
}

\date{Received: date / Accepted: date}

\maketitle

\begin{abstract}

For the goal of automated design of high-performance deep  convolutional neural networks (CNNs), Neural Architecture Search (NAS) methodology is becoming increasingly important for both academia and industries.
Due to the costly stochastic gradient descent (SGD) training of CNNs for performance evaluation, most existing NAS methods are computationally expensive for real-world deployments. To address this issue, we first introduce a new performance estimation metric, named Random-Weight Evaluation (RWE) to quantify the quality of CNNs in a cost-efficient manner. Instead of fully training the entire CNN, the RWE only trains its last layer and leaves the remainders with randomly initialized weights, which results in a single network evaluation in seconds.
Second, a complexity metric is adopted for multi-objective NAS to balance the model size and performance. Overall, our proposed method obtains a set of efficient models with state-of-the-art performance in two real-world search spaces. Then the results obtained on the CIFAR-10 dataset are transferred to the ImageNet dataset to validate the practicality of the proposed algorithm. Moreover, ablation studies on NAS-Bench-301 datasets reveal the effectiveness of the proposed RWE in estimating the performance compared with existing methods.

\end{abstract}

\input{intro}
\input{related-work}
\input{method}

\input{results}

\section{Conclusion}
This paper proposed a flexible performance metric Random-Weight Evaluation (RWE) to rapidly estimate the performance of CNNs. Inspired by the expressive power of randomly initialized convolution filters, RWE only trains the last classification layer and leaving the backbone with randomly initialized weights. As a result, RWE achieves a reliable estimation of architectures in seconds. We further integrated RWE with an evolutionary multi-objective algorithm, adopting the complexity metric as the second objective. The experimental results showed that our algorithm achieved a set of efficient networks with state-of-the-art performance on both micro and macro search spaces. The resulted architecture with 350M FLOPs achieved 2.98\% Top-1 error in CIFAR-10 and 27.6\% Top-1 error in ImageNet after transferring. Also, the careful ablation studies experiments on different performance metrics and initialization methods indicated the effectiveness of the proposed algorithm.

\begin{acknowledgements}
This work was supported by the National Natural Science Foundation of China (No. 61903178, 61906081, and U20A20306), the Shenzhen Science and Technology Program (No. RCBS20200714114817264), the Program for Guangdong Introducing Innovative and Entrepreneurial Teams (No. 2017ZT07X386), the Shenzhen Peacock Plan (No. KQTD2016112514355531), and the Program for University Key Laboratory of Guangdong Province (No. 2017KSYS008).
\end{acknowledgements}

%
\section*{Conflict of interest}
The authors declare that they have no conflict of interest.

\bibliographystyle{spmpsci}      
\bibliography{wann}   

\end{document}

%% file: intro.tex
\section{Introduction}

In recent years, deep convolutional neural networks (CNNs) have been widely studied and achieved astonishing performance in different computer vision tasks. 
One crucial component among these studies is the design of dedicated architectures of neural networks, which significantly affects the performance and generalization ability of CNNs among various tasks \cite{krizhevsky2012imagenet,he2016deep,muxconv}. 
Along with the architectural milestones, from the original AlexNet \cite{krizhevsky2012imagenet} to the ResNet \cite{he2016deep}, the performance of CNNs across extensive datasets and tasks keeps boosting. 
However, it still takes researchers enormous works to achieve these architectural advancements via trial-and-error tuning manually. 
Therefore, Neural Architecture Search (NAS) has emerged as an alternative way to design CNN in an automated manner. 
Although NAS alleviates the laborious experiments by researchers, existing NAS algorithms still suffer from numerous computational overheads, leading to challenges in the real-world deployment \cite{Zoph2018,Real2019}.

The expensive evaluations of the performance of architectures contribute to dominant computational consumption in the NAS algorithm.
Usually, a brute-force training of a network can cost days to weeks on a single GPU, varying from simple to complex datasets and tasks. 
Therefore, several approaches have been proposed to approximate the true performance with fewer computational costs and, as a result, fewer fidelities. 
These works can be roughly divided into three categories.

The first category includes methods that reduce training budgets via decreasing the network sizes (e.g., the number of layers and channels), which are widely adopted in early NAS works \cite{Zoph2018,Real2019,he2021efficient}.
Nevertheless, their effectiveness has not been systematically researched until recently \cite{zela2018,Zhou_2020_CVPR}, demonstrating that their effectiveness can be limited under inappropriate parameter settings. 
Moreover, these methods are computationally expensive due to the thorough training for every single network. 
Finally, these methods require that the CNN architectures in search spaces should be modular, i.e., the networks are constructed by repeatedly stacking modular blocks.
For instance, several state-of-the-art search spaces in \cite{Liu2018a,lu2020nsganetv2} do not follow the constraints above, and the extension of these methods to new search spaces is not trivial. 

The second category is often known as the supernet-based method, which intends to avoid training every architecture from scratch \cite{tan2020relativenas,lu2020nsganetv2,lu2021neural}. 
This technique typically decouples NAS into two main stages to share weights during searching. 
In the first stage, it constructs a supernet that contains all possible architectures in the search space, such that each architecture becomes a subnet of the supernet. 
In the second stage, the search process begins, and each architecture inherits the weights from the supernet, and thus the evaluation of each architecture becomes a simple inference on the valid set. 
Despite the fact that this technique can speed up the searching process, the construction of the supernet could be more time-consuming than a complete search \cite{cai2020once}. 
Besides, the search spaces require substantial modifications to accommodate the construction of the supernet \cite{Liu2019}.

The third category consists of several studies known as the zero-cost proxies \cite{abdelfattah2021zerocost,mellor2020neural}, which estimate the performance with a few mini-batches of forward/backward propagation. 
To be more specific, this category analyzes information such as gradients, activations, or magnitudes of parameters to achieve estimations for reducing the computational cost drastically.
Notably, most of these techniques attempt to validate their effectiveness on several NAS-Bench datasets \cite{siems2020bench,dong2020nasbench201} which are public architecture datasets constructed by exhaustively evaluating search spaces. 
Nevertheless, they may perform well only on certain NAS-Bench datasets \cite{abdelfattah2021zerocost}, or they are not validated on real-world search spaces \cite{mellor2020neural}.

On top of the performance estimation methods, a branch of works named predictor-based NAS \cite{lu2020nsganetv2,lu2021neural} has been proposed to further improve the sampling efficiency. In these works, a regression model, i.e., a performance predictor, would be trained to fit the mapping from the architectures to the corresponding performance. After the establishment of the predictor, the estimated performance in the searching stage is achieved by the evaluations of the predictor instead of the expensive estimation methods, which improves the sampling efficiency of NAS. The predictor can be built upon different performance estimation methods, e.g., based on the training with reduced budgets \cite{8744404,wen2020neural} and the evaluations of the supernet \cite{lu2020nsganetv2,lu2021neural}. Also, several works explore different encoding methods for the architectures \cite{Sun2021,ning2020generic,yan2020does} and different machine learning models as the predictor \cite{ning2020generic,8744404,wen2020neural}. 
 
In this work, we propose a Random-Weight Evaluation (RWE) approach. Comparing to the existing methods, it is less expensive, more flexible, and the effectiveness is validated more solid. 
In detail, by training the last classification layer only and keeping others with randomly initialized weights, RWE saves orders of magnitudes computational resources comparing with conventional methods. 
At the same time, RWE is conceptually flexible with any search space, and it does not need any modifications to the search space. 
Moreover, the effectiveness of RWE is validated by the searching on two modern real-world search spaces and some ablation studies on the NAS-Bench-301 dataset. 
We briefly summarize our main contributions below:
\begin{itemize}
\item We propose a novel performance estimation metric, namely RWE, for efficiently quantifying the quality of CNNs. 
RWE is highly efficient in computational costs compared with conventional methods, which reduces the wall-clock evaluation time from hours to seconds. 
Extensive experiments on both real-world search spaces and benchmark search space NAS-Bench further validate the effectiveness of RWE.
\vspace{.2cm}
\item Paired with a multi-objective evolutionary algorithm, our RWE based NAS algorithm can achieve a set of efficient networks in one run, where both the performance and efficiency of models are considered. 
For instance, the proposed algorithm achieves state-of-the-art performance on the CIFAR-10 dataset, resulting in the networks from the largest, with 2.98\% Top-1 error and 1.5M parameters, to the smallest, with 4.05\% Top-1 error and 0.9M parameters. The experiments of transferability on ImageNet further demonstrate the competitiveness of our method.
With such competitive performance, the whole searching procedure only costs less than two hours on a single GPU card, making the algorithm highly practical in handling real-world applications.

\end{itemize}

The rest of this paper is organized as follows. 
In Section 2, some related work on multi-objective NAS algorithms and the expressive power of randomly initialized convolution filters is introduced.
Then we present our proposed approach in Section 3, including the detailed random-Weight evaluation, search strategy, and search space and encoding.
Comparative studies are shown in Section 4, and the conclusions are drawn in Section 5.

%% file: related-work.tex
\section{Related Works}
In this section, we briefly discuss two topics related to the technicalities of our approach, i.e., multi-objective NAS and randomly initialized convolution filters.

\subsection{Multi-Objective NAS} 
Single-objective optimization algorithms have dominated the early researches in NAS \cite{Zoph2018,Liu2019,Real2019}, which mainly propose architectures to maximize their performance on certain datasets and tasks. 
Though NAS algorithms have shown their practicality in solving benchmark tasks, they cannot meet the demands from deployment scenarios varying from GPU servers to edge devices \cite{Howard2017}. 
Thus, NAS algorithms are expected to balance between multiple conflicting objectives, such as inference latency, memory footprint, and power consumption. 
Though recent attempts often convert multiple competing objectives into a single objective in a weighted sum manner \cite{Tan2019,Cai2019}, they may miss the global optima of the problem. As a result, multiple runs of the algorithm could be required in real-world applications, due to the difficulty of deciding the best coefficient for weighted sum. Also, the search strategies these works adopted are primarily based on gradient-based methods or reinforcement learning, and they cannot approximate the Pareto front in a single run.

There are also several works that adopt multi-objective evolutionary  algorithms as search strategies for NAS \cite{Lu2019,lu2020nsganetv2,lu2021neural}. 
The population-based strategies introduce the natural parallelism, which increases the practicality in large-scale applications, and the conflicting nature of multiple objectives is helpful to enhance the diversity of the population. 
Most of them aim to tradeoff between the performance and the complexity of networks \cite{Lu2019,lu2020nsganetv2}, while some other works temp to exploit the performance among different datasets, similar to the concepts in multi-task learning \cite{lu2021neural}. 
Following successful practices, we adopt a classic multi-objective evolutionary algorithm, namely NSGA-II \cite{deb2002fast}.
We aim to achieve a set of efficient architectures in one run, where the proposed performance metric and a complexity metric FLOPs are two conflicting objectives to be optimized.

\subsection{Expressive Power of Randomly Initialized Convolution Filters}
The RWE is surprisingly powerful, inspired by the fact that the convolution filters are in terms of extracting the features for input images, even with randomly initialized weights \cite{Jarrett2009}. 
It is indicated in \cite{Jarrett2009,wann2019} that, with proper architecture, the convolution filters with randomly initialized weights can be as competitive as the ones with fully trained weights on both visual and control tasks. 
Also, it is validated that the structure itself can introduce prior knowledge to be capable of capturing the features for visual tasks~\cite{Adebayo2018,ulyanov2018deep}. 
Similarly, the local binary convolutional neural network achieves comparable performance to CNNs with fully trained convolution filters by learning a linear combination of randomly initialized convolution filters \cite{lbcnn}. 

Some early works in the literature conceptually explore the potential of estimating the performance of networks from randomly initialized weights. In detail, Sax \emph{et al.} mathematically proved that the convolutional ﬁlter with random weights still has its key functionality, which is frequency selectivity and translation invariance. These characteristics are utilized to rank shallow neural networks with different network configurations \cite{Saxe2011}. Rosenfeld and Tsotsos successfully predict the performance ranking on several widely used CNN architectures by only training a fraction of weights in convolutional filters \cite{Rosenfeld2019}. 

Although previous works show the potential of randomly initialized convolution filters, those methods are not scalable to real-world applications. In this work, we randomly initialized and freeze the weights in convolutional kernels in CNN, only training for the last classification layer. Using the predictive performance after doing so as a performance metric, we demonstrate the scalability of our approach on complex datasets and modern CNN search spaces that contains deep yet powerful CNNs.

%% file: method.tex
\section{Proposed Approach}
The Multi-objective NAS problem for a target dataset $\mathcal{D} = $ $\{ \mathcal{D}_{trn}, \mathcal{D}_{vld}, \mathcal{D}_{tst} \}$  can be formulated as the following bilevel optimization problem  \cite{lu2020},
\newcommand{\argmin}{\operatornamewithlimits{argmin}}
\begin{mini*}|l|
{\boldsymbol{\alpha}}{f_1(\boldsymbol{\alpha};\boldsymbol{w^*}(\boldsymbol{\alpha})),f_2(\boldsymbol{\alpha}),...,f_m(\boldsymbol{\alpha})}
{}{}
\addConstraint{\boldsymbol{w^*}(\boldsymbol{\alpha}) \in \argmin_{\boldsymbol{w}} \mathcal{L}(\boldsymbol{w};\boldsymbol{\alpha})}
\addConstraint{\boldsymbol{\alpha} \in \Omega_\alpha, \hspace{1em}\boldsymbol{w} \in \Omega_w}
\end{mini*}
where the upper lever variable $\boldsymbol{\alpha}$ defines an architecture in the search space $\Omega_\alpha$, and the lower level variable $\boldsymbol{w}(\boldsymbol{\alpha})$ represents the corresponding weights. $\mathcal{L}(\boldsymbol{w};\boldsymbol{\alpha})$ is the loss function on the $\mathcal{D}_{trn}$ for the architecture $\boldsymbol{\alpha}$ with weights $\boldsymbol{w}$. The first objective $f_1$ represents the classification error on $ \mathcal{D}_{vld}$, which depends on both architectures and weights. Other objectives $f_2,...,f_m$ only depend on architectures, such as the number of parameters, floating-point operations (FLOPs), latencies, etc.

In our approach, we simplify the complex bilevel optimization by using the proposed performance metric RWE as a proxy of $f_1$. In addition, we adopt the complexity metric FLOPs as the second objective $f_2$ to optimize. As a result, the multi-objective formulation of this work becomes
\begin{mini*}|l|
{\boldsymbol{\alpha}}{\text{RWE}(\boldsymbol{\alpha}), \text{FLOPs}(\boldsymbol{\alpha})}
{}{}
\addConstraint{\boldsymbol{\alpha} \in \Omega_\alpha},
\end{mini*}
where RWE and FLOPs represent the values of these metrics with respect to architecture $\alpha$.

\subsection{Random-Weight Evaluation}

\begin{algorithm}[!hbt]   
    \SetAlgoLined
    \SetKwInOut{Input}{Input}\SetKwInOut{Output}{Output}
    \Input{An architecture $\boldsymbol{\alpha}$, a training and validation dataset \begin{math}\mathcal{D}_{trn}\text{,  }\mathcal{D}_{vld}\end{math}}
    \Output{The performance metric \emph{RWE} of $\boldsymbol{\alpha}$.}
    
    \emph{net} $\leftarrow$ Decode the architecture $\boldsymbol{\alpha}$ into CNN backbone;
    
    Randomly initialize the \emph{net} and a linear classifier \emph{clsfr};
    
    Freeze the weights of \emph{net} throughout the whole algorithm;
    
    \emph{features} $\leftarrow$ Infer $\mathcal{D}_{trn}$ on the \emph{net};
        
    Train the linear classifier \emph{clsfr} with the \emph{features} as input;
    
    \ForEach{image, target $\in$ $\mathcal{D}_{vld}$}{
    
        Infer \emph{image} on \emph{net} with \emph{clsfr};
        
        \emph{prediction} $\leftarrow $ the label approved by \emph{clsfr};
        
        Compare the \emph{prediction} with \emph{target} and record the result;
    }
    
    \emph{RWE} $\leftarrow$ Calculate the classification error rate;
    
    \Return{RWE;}
 \caption{Random-Weight Evaluation}
    
    \label{algorithm1}
    
\end{algorithm}

As mentioned in Section 2.2, randomly initialized convolution filters are surprisingly powerful in extracting the features from images, due to the frequency selectivity and translation invariance preserved with random weights \cite{Saxe2011}. Inspired by this amazing characteristic, this work attempts to judge the quality of the architectures based on the ability of architectures with random weights to extract ``good'' features. And we quantify the quality of the features by training a linear classifier that takes these features as input and calculates the classification error for that classifier.

We detail the proposed performance metric Random-Weight Evaluation (RWE) as following. The overall procedure is shown in the Algorithm 1. First, we decode the encoding of a candidate architecture $\alpha$ into a CNN backbone \emph{net}, which refers to all layers before the last classification layer. Second, we initialize the \emph{net} and a linear classifier \emph{clsfr} with random weights, the latter of which acts as the last classification layer in a complete CNN and its structure is identical for all candidate CNNs in the search space. Here, a modified version of the \emph{Kaiming initialization} \cite{He} is adopted to initialize the \emph{net} (default setting in PyTorch).  The weights in the backbone part will keep frozen throughout the algorithm. Third, we infer the training set $D_{trn}$ on \emph{net} and utilize the output \emph{feature} to train \emph{clsfr}. Finally, after assembling \emph{net} and trained \emph{clsfr} into a complete CNN, this CNN gets tested on the validation set $D_{vld}$, the output error rate of which becomes the value of RWE.

\subsection{Search Strategy}
We adopt a classic multi-objective evolutionary algorithm NSGA-II \cite{deb2002fast} in our approach, where the searching process is detailed as below.

First, we randomly initialize the population, the individual of which get evaluated with RWE and FLOPs as two objectives. Second, we apply the binary tournament selection to select the parents for offspring. Third, the two-point crossover and the polynomial mutation are applied to generate the offspring, followed by the evaluations of offspring. Finally, we apply the environment selection based on the nondominated sorting and the crowding distance \cite{deb2002fast}, and the process is repeated until reaching the max generation. 

\subsection{Search Space and Encoding}

\begin{figure*}[!hbt]
     \centering
     \includegraphics[width=0.85\linewidth]{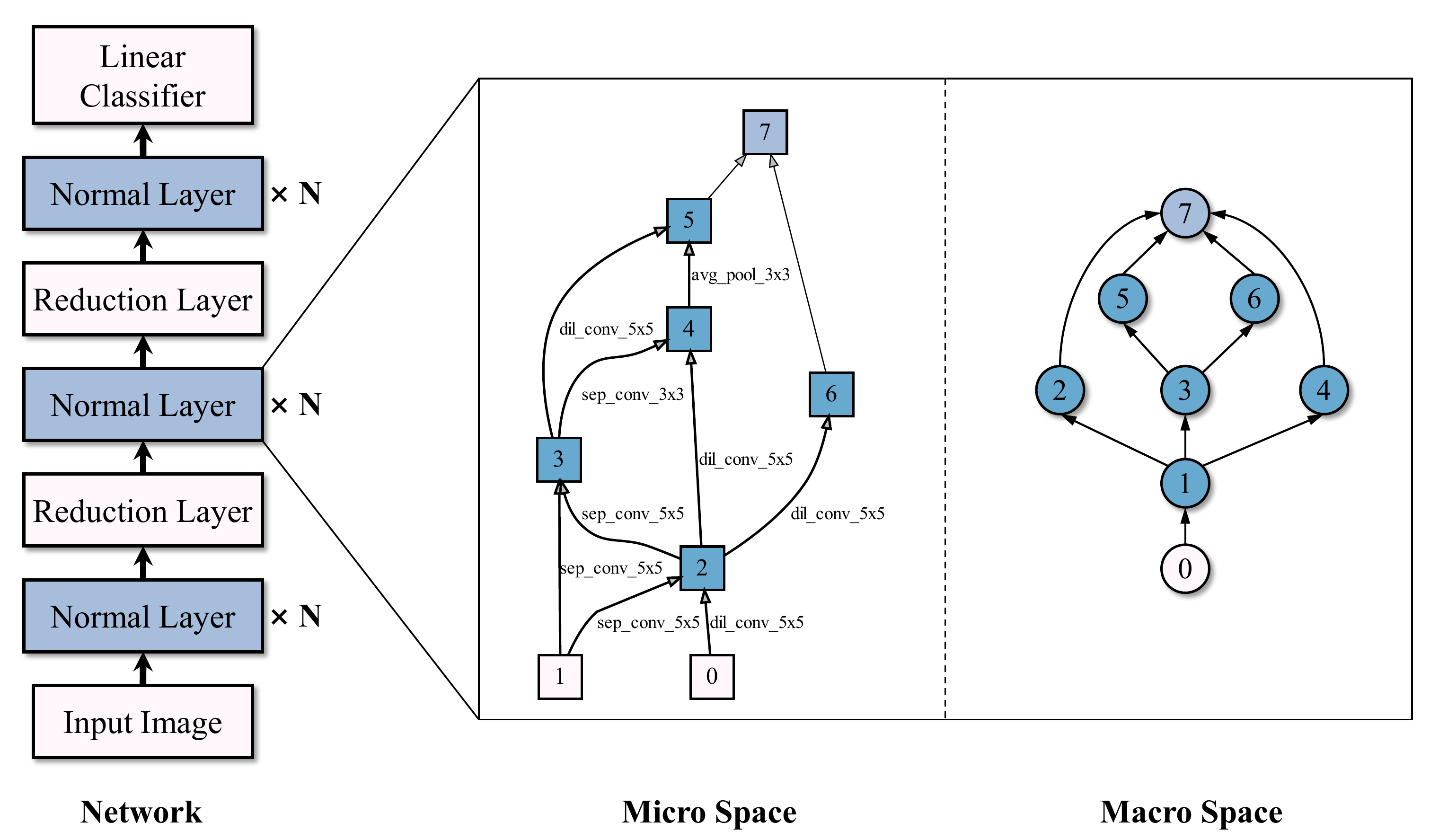}
    \caption{The micro \cite{Zoph2018} and macro \cite{Xie2017} search spaces adopted in our approach. LEFT: Overall network architecture. MIDDLE and RIGHT: Design of layers in micro and macro search space.}
    \label{searchspace}
\end{figure*}

Our proposed RWE is conceptually flexible and can be applied to any search space. To validate the effectiveness of our algorithm on the real-world application, we experiment with two modern search spaces, the micro \cite{Zoph2018} and macro \cite{Xie2017} search spaces, in our approach. As shown in Fig. \ref{searchspace} LEFT, these two search spaces are modular search spaces, in which two kinds of layers, the normal and reduction layers, are repeatedly stacked, forming the complete CNN. The former kind of layers keeps the resolution and the number of channels for input images, while the latter halves the resolution and doubles the number of channels. The main difference between the micro and macro spaces is the design of each layer and the way to stack them into a complete CNN.

\vspace{0.5em}
\noindent\textbf{Micro Search Space:}
In the micro search space \cite{Zoph2018}, we search for both the normal and reduction layers, named the normal and reduction cells. All the normal cells share the same architecture in a CNN, in which the weights are different though, and the case for the reduction cells is the same. Typically, we scale networks using different repeating number ($\mathbf{N}$) in searching and validation stages. The normal and reduction cells share the same template, except for the stride size in operators. In each kind of cell, we search for both of the connections between nodes and the operation applied on each connection, as shown in Fig. \ref{searchspace} MIDDLE.

\noindent\textbf{Macro Search Space:}
In the macro search space \cite{Xie2017}, we search for only the normal layers, leaving the predefined reduction layers identical. Each normal layer in the macro search space is searched independently, where the repeating number in a phase ($\mathbf{N}$) is equal to one. In the normal layers, only the connection patterns get searched, and the operation in each node is a predefined sequential operator, including convolution operators, batch normalization layers, and activation functions. Fig. \ref{searchspace} RIGHT shows an example for candidate connection patterns.

%% file: results.tex
\section{Experimental Results}
In this section, we first present the searching results of our proposed NAS algorithm on the micro and macro search spaces for a modern classification dataset CIFAR-10 \cite{krizhevsky2009learning}. Then, the ablation studies on NAS-Bench-301 \cite{siems2020bench} demonstrate the effectiveness of our evaluation method and the rationality of some design choices. Finally, the experiment on ImageNet \cite{imagenet_cvpr09}, which is one of the most challenging classification benchmarks, shows the transferability of our architectures and illustrates the practicality for real-world applications.

\subsection{Searching on CIFAR-10\label{section4.2}}
In our approach, we search on a modern classification dataset CIFAR-10 \cite{krizhevsky2009learning}, which contains ten categories and 60K $32\times 32$ images. Conventionally, the dataset split into a training set with 50K images and a test set with 10K images. Following common settings in NAS algorithms \cite{Zoph2018,Real2019,Lu2019}, we further split the training set (80\%-20\%) in the searching stage to create the training and validation sets.
 
Here we introduce the detailed implementation and parameter settings in our NAS algorithm. In the searching stage, the population size is set to 20 and the max generation is set to 30. For RWE, the architectures in the micro search space have 10 initial channels and 5 layers, and the architectures in the macro search space have 32 initial channels. Also, due to the randomness introduced in RWE, we adopt assemble learning technique \cite{Hansen1990} in the training of the linear classifier to stabilize the results. Specifically, there are five classifiers to be trained, each of which is only exposed to 4/5 features. We only introduce the normalization techniques in the preprocessing of the input images, without the data augmentation techniques introducing the randomness. SGD optimizer with an initial learning rate of 0.25 and a momentum of 0.9 is adopted, where the cosine annealing schedule \cite{loshchilov2016sgdr} decays the learning rate to zero gradually. The batch size is set to 512 and the training iterations conduct for 30 epochs. The average CPU time for a single evaluation is approximately 10 seconds with a single Nvidia 2080Ti.

      

\begin{table*}[!hbt]
\centering
\caption{The results of the proposed algorithm and other state-of-the-art methods on CIFAR-10. $^{\Updownarrow}$ denotes the results achieved by the same training setting with ours and reported in \cite{Lu2019}. $^{\dagger}$ denotes the work that adopts the regularization technique cutout \cite{devries2017improved}.} 
\resizebox{0.98\textwidth}{!}{%
\begin{tabular}{@{\hspace{2mm}}lccccc@{\hspace{2mm}}}
\toprule
Architecture &
  \multicolumn{1}{c}{\begin{tabular}[c]{@{\hspace{2mm}}c@{}}Test Error\\ (\%)\end{tabular}} &
  \multicolumn{1}{c}{\begin{tabular}[c]{@{\hspace{2mm}}c@{}}Params\\ (M)\end{tabular}} &
  \multicolumn{1}{c}{\begin{tabular}[c]{@{\hspace{2mm}}c@{}}FLOPs\\ (M)\end{tabular}} &
  \multicolumn{1}{c}{\begin{tabular}[c]{@{\hspace{2mm}}c@{}}Search Cost\\ (GPU days)\end{tabular}} &
  \hspace{2mm} Search Method \\ \hline
Wide ResNet \cite{Zagoruyko2016}       & 4.17 & 36.5 & -   & -     & manual         \\
DenseNet-BC \cite{Huang2017}             & 3.47 & 25.6 & -   & -     & manual         \\ \midrule
BlockQNN$^{\dagger}$  \cite{Zhong2020}                    & 3.54 & 39.8 & -   & 96    & RL \\ 
SNAS$^{\dagger}$ \cite{Xie2019}    & 3.10 & 2.3  & -   & 1.5   & gradient  \\
NASNet-A$^{\dagger\Updownarrow}$  \cite{Zoph2018}                 & 2.91 & 3.2  & 532 & 2,000 & RL             \\
DARTS$^{\dagger\Updownarrow}$  \cite{Liu2019}         & 2.76 & 3.3  & 547 & 4     & gradient \\

          \midrule
 NSGA-Net$^{\Updownarrow}$ + macro space  \cite{Lu2019}                 & 3.85 & 3.3  & 1290 & 8     & evolution      \\
\textbf{Macro-L$^{\dagger}$} & $\boldsymbol{4.27}$ & $\boldsymbol{2.79}$ & $\boldsymbol{1074}$ & $\boldsymbol{0.14}$  & evolution\\ 
 \midrule
AE-CNN + E2EPP \cite{8744404} & 5.30 & 4.3  & - & 7     & evolution      \\
Hier. Evolution \cite{Liu2018a}               & 3.75 & 15.7 & -   & 300   & evolution     \\ 
AmoebaNet-A$^{\dagger\Updownarrow}$  \cite{Real2019}              & 2.77 & 3.3  & 533 & 3,150 & evolution      \\
NSGA-Net$^{\dagger\Updownarrow}$  \cite{Lu2019}                 & 2.75 & 3.3  & 535 & 4     & evolution      \\

 \midrule
\textbf{Micro-S$^{\dagger}$} & $\boldsymbol{4.05}$ & $\boldsymbol{0.9}$ & $\boldsymbol{203}$ & $\boldsymbol{0.05}$  & evolution\\
\textbf{Micro-M$^{\dagger}$} & $\boldsymbol{3.37}$ & $\boldsymbol{1.2}$ & $\boldsymbol{249}$ & $\boldsymbol{0.05}$  & evolution\\
\textbf{Micro-L$^{\dagger}$} & $\boldsymbol{2.98}$ & $\boldsymbol{1.5}$ & $\boldsymbol{340}$ & $\boldsymbol{0.05}$  & evolution\\

\bottomrule
\end{tabular}}
\label{cifar10res}
\end{table*}

\begin{figure*}
    \centering
    \includegraphics[width=\textwidth]{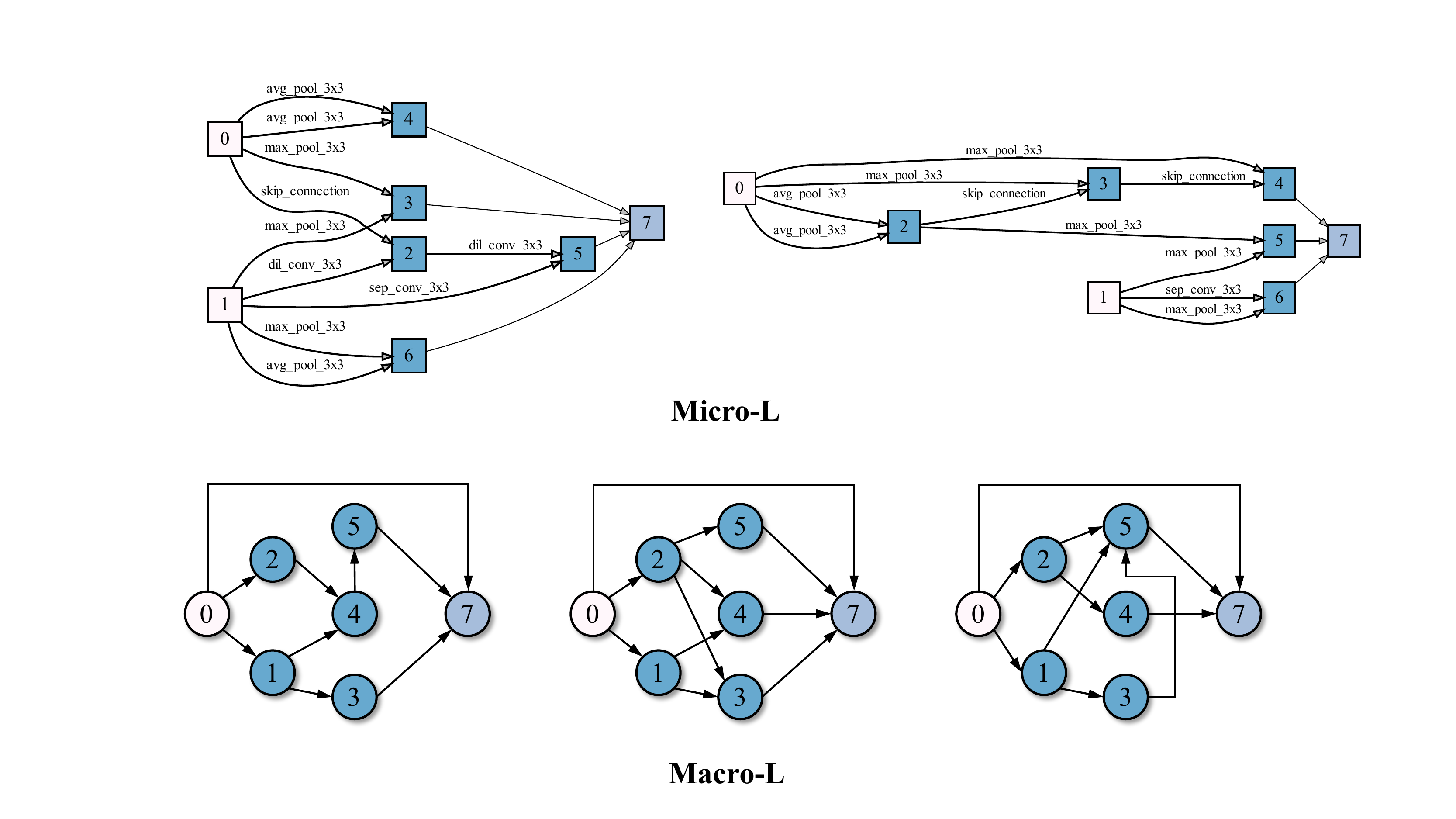}
    \caption{The visualization of \textbf{Micro-L} and \textbf{Macro-L} architecture.
    }
    \label{visualization}
\end{figure*}

For the validation stage, we scale the architectures to the deployment settings, where the number of training epochs, layers, and channels increases. The architectures from the final Pareto front are selected to be trained from scratch, where the number of layers and initial channels is set to 20 and 34 for the micro search space, and the number of channels is set to 128 in all layers in the macro search space. We use the same SGD optimizer as the one in the training stage, except the initial learning rate is set to 0.025. The selected architectures are trained for 600 epochs with a batch size of 96. Also, the regularization techniques cutout \cite{devries2017improved} and scheduled path dropout \cite{Zoph2018} is introduced, where the cutout length and the drop out rate are set to 16 and 0.2. The settings are the same with state-of-the-art algorithms for a fair comparison \cite{Lu2019}.

The results of validation and the comparison to other state-of-the-art architecture are shown in Table. \ref{cifar10res}. The representative architectures from the final Pareto front get compared to both hand-crafted and search-based architectures. In the experiments with the micro search space, the architecture with the lowest error rate (\textbf{Micro-L}) in our approach achieves a 2.98\% Top-1 error rate with 340M FLOPs. With fewer FLOPs, it has competitive performance with state-of-the-art architectures. Also, \textbf{Micro-M, Micro-S} shows a different tradeoff between performance and complexity. Similar to the micro search space, the chosen architecture in the macro search space (\textbf{Macro-L}) has competitive performance and fewer FLOPs comparting to the state-of-the-art. The visualization of the detailed structures of \textbf{Micro-L} in the micro space and \textbf{Macro-L} in the macro space are shown in Fig. \ref{visualization}.

\subsection{Effectiveness of Random-Weight Evaluation}
\label{section:effectiveness}

\begin{figure}
    \centering
    \includegraphics[width=\linewidth]{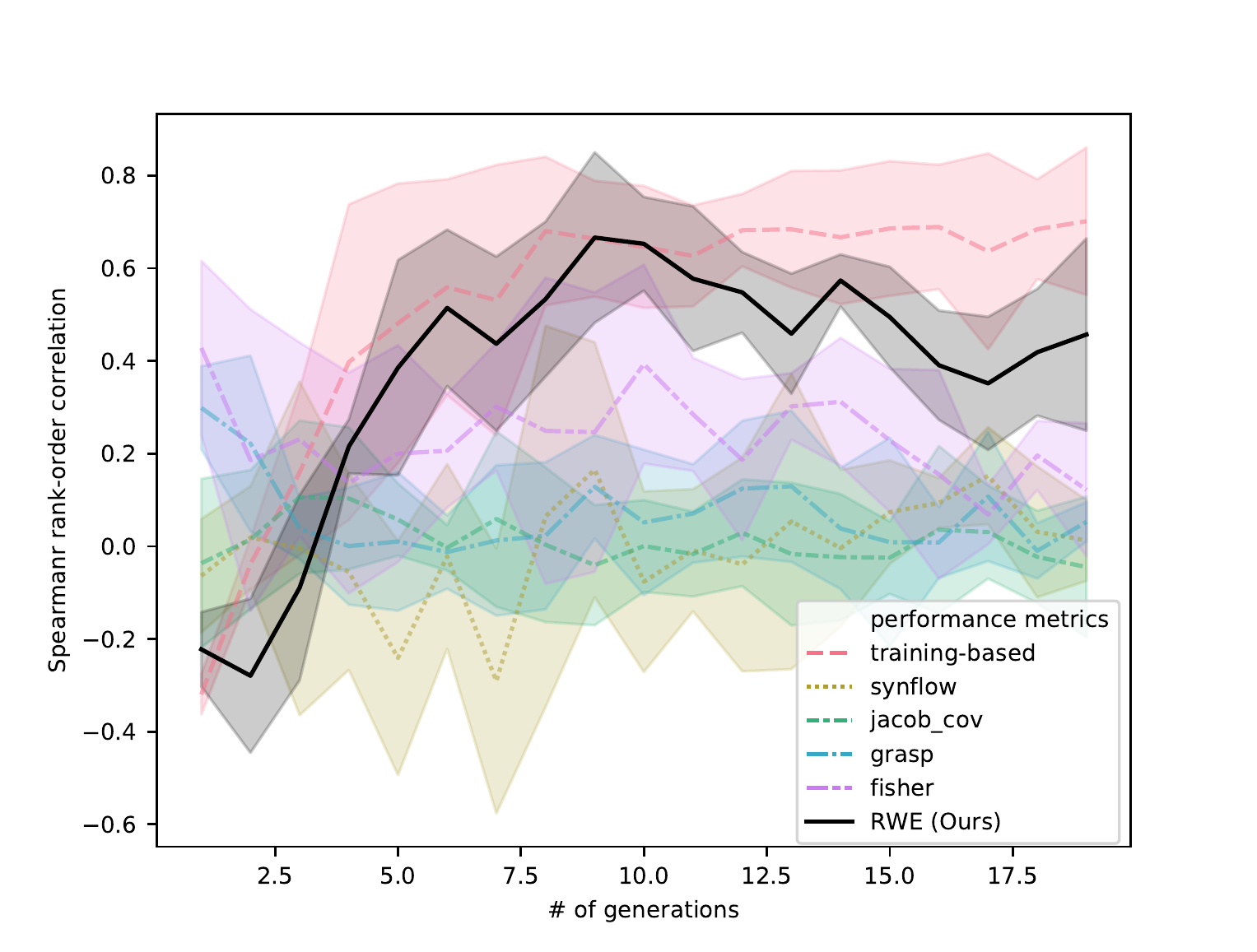}
    \caption{The Spearman correlation coefficient for different performance metrics in searching on NAS-Bench-301.}
    \label{corr_comp}
\end{figure}

\begin{figure}
    \centering
    \includegraphics[width=\linewidth]{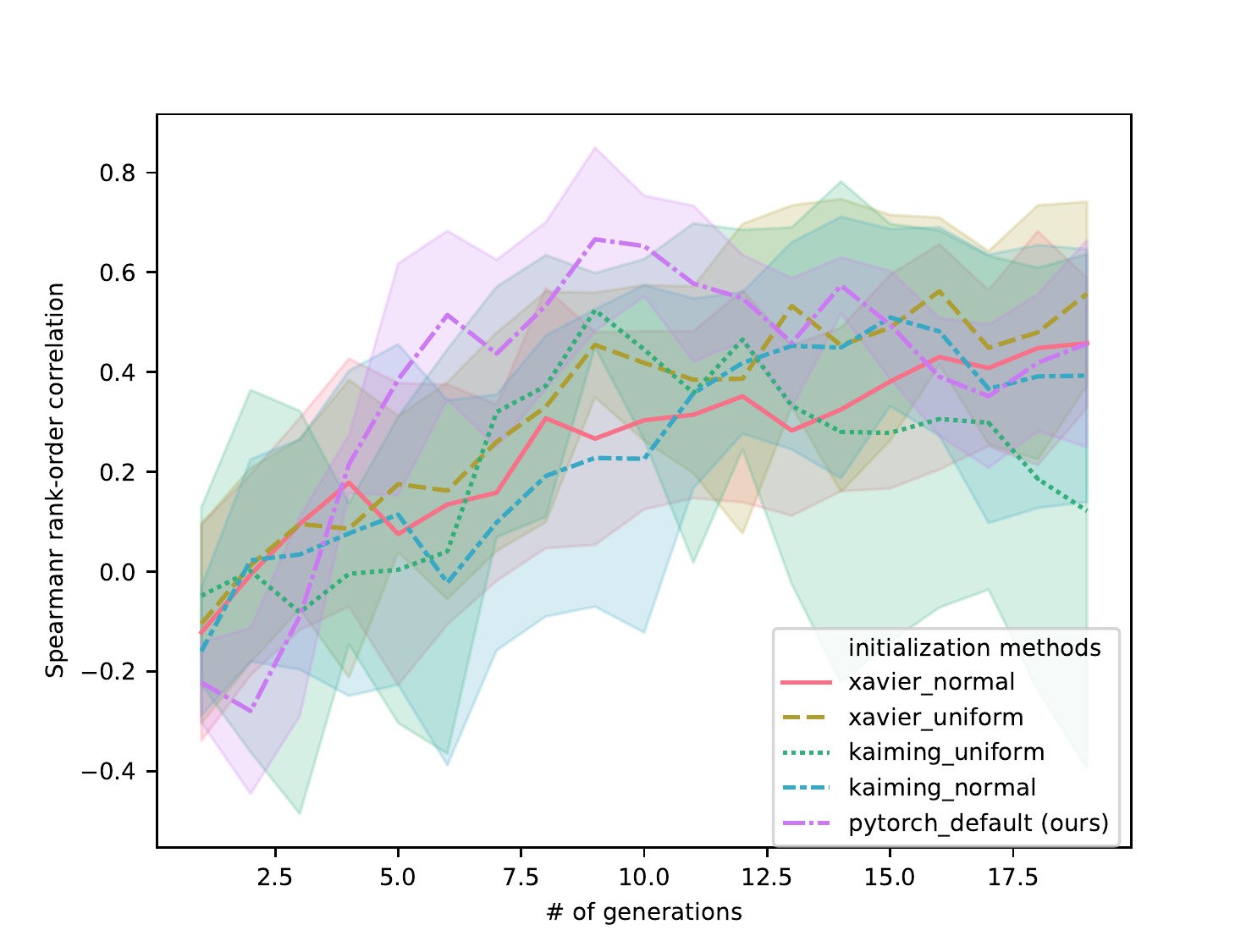}
    \caption{The Spearman correlation coefficient for different initialization method in searching on NAS-Bench-301.}
    \label{corr_init}
\end{figure}

To demonstrate the effectiveness of RWE, we conduct experiments on NAS-Bench-301 dataset \cite{siems2020bench}. The dataset is constructed by the surrogate model trained by sampled architectures in search space, such that it covers the whole search space to help researchers to analyze their NAS algorithms. While other NAS-Bench datasets construct a toy search space for the convenience of studies \cite{dong2020nasbench201}, NAS-Bench-301 covers a real-world search space, which is the micro search space adopted in our work. Thus, the ablation studies based on NAS-Bench-301 examine the behavior of our algorithm during the searching stage.

We evaluate the effectiveness of estimation strategies by calculating the Spearman correlation coefficient between the estimation and queried performance from NAS-Bench-301. The target individuals are from the union of the population of each generation and their offspring. The Spearman correlation coefficient, which ranged from $[-1,1]$, is a nonparametric measure of rank correlation. The higher coefficient is, the ranking of two variables has a more similar rank to each other. The idea of experiment settings is that, during the evolutionary algorithm, the only phase depending on the estimation strategy is the mating and survival selection, which happen in the union mentioned above. The higher the correlation coefficient is, the more reliable the estimation strategy is. Thus, the algorithm has more chances to choose good candidates from a set of architectures. 

In the following experiments, we use the same search strategy as introduced in Section 3.2 and conduct the experiments in NAS-Bench-301 with 20 generations. The search space in NAS-Bench-301 is a subset of the Micro Space, where identical connections to a single node are not allowed in NAS-Bench-301. As a result, we add a fix operation in the search strategy, which randomly chooses another connection to avoid duplication when an invalid architecture is produced. Also, the results present the mean and the standard variation of five independent trials with different random seeds.

We first compare our estimation strategy RWE with the zero-cost proxies \cite{abdelfattah2021zerocost,mellor2020neural} and the training-based evaluation method \cite{Zoph2018,Zhou_2020_CVPR}. For the zero-cost proxies, we choose the representative performance metrics \emph{synflow} , \emph{grasp}, and \emph{fisher} from \cite{abdelfattah2021zerocost}, and \emph{jacob\_conv} from \cite{mellor2020neural}. For the training-based method, we train the network for 10 epochs with the number initial channels and layers of 16 and 8. As shown in Fig. \ref{corr_comp}, the proposed RWE outperforms all zero-cost proxies after the initial stage of searching, ending up with a similar accuracy with the training-based method. The experiment shows the effectiveness of RWE is competitive to the one of training-based method while having much fewer computational overheads. Paired with the searching in the micro and macro spaces, it further shows that RWE performs well in the real-world search spaces.

We then investigate the effects of different initialization methods applied in our approach. The method we adopt in this paper is the default one in PyTorch, and we examine other four representative initialization methods, which are known as \emph{Kaiming normal (uniform) initialization} \cite{He} and \emph{Xavier normal (uniform) initialization}  \cite{glorot2010understanding}. As shown in Fig. \ref{corr_init}, the initialization methods have minor impacts on the effectiveness of RWE, as we observe no significant different behaviors. The experiment shows that our approach is robust to different initialization methods.

\subsection{Transferring to ImageNet}

\begin{table}[!hbt]
\caption{The results of the proposed algorithm and other state-of-the-art methods on ImageNet. $^{\Uparrow}$ denotes the methods that first get searched on CIFAR-10 and then get transferred to ImageNet.}
	\centering
	\begin{tabular}{lcccc}
\toprule
\multicolumn{1}{l}{\multirow{2}{*}{Architecture}} &
  \multicolumn{2}{c}{\begin{tabular}[c]{@{}c@{}}Test Error (\%)\end{tabular}} &
  \multicolumn{1}{c}{\multirow{2}{*}{\begin{tabular}[c]{@{}c@{}}Params\\ (M)\end{tabular}}} &
  \multicolumn{1}{c}{\multirow{2}{*}{\begin{tabular}[c]{@{}c@{}}FLOPs\\ (M)\end{tabular}}} \\ \cline{2-3}
\multicolumn{1}{l}{}                          & top-1 & top-5 & \multicolumn{1}{c}{} & \multicolumn{1}{c}{} \\ \midrule
MobileNetV1 \cite{Howard2017}                          & 31.6  & -     & 2.6                  & 325                  \\
InceptionV1 \cite{Szegedy2015}                                   & 30.2  & 10.1  & 6.6                  & 1448                 \\
ShuffleNetV1 \cite{Zhang2018}                          & 28.5  & -     & 3.4                  & 292                  \\
VGG \cite{simonyan2014very}  &  28.5  & 9.9   & 138     & - \\
MobileNetV2 \cite{Sandler2018}                          & 28.0  & 9.0   & 3.4                  & 300                  \\
ShuffleNetV2 1.5$\times$ \cite{ma2018shufflenet}   &  27.4  &  -  &  -     & 299 \\
\midrule
NASNet-C $^{\Uparrow}$ \cite{Zoph2018} & 27.5  & 9.0   & 4.9                  & 558                  \\
SNAS $^{\Uparrow}$ \cite{Xie2019}          & 27.3  & 9.2   & 4.3                  & 533                  \\ 
EffPNet $^{\Uparrow}$ \cite{9349967}          &  27.0  & 9.25   & 2.5     & - \\
DARTS $^{\Uparrow}$ \cite{Liu2019}          &  26.7  &  8.7   & 4.7    & 574 \\
AmoebaNet-B $^{\Uparrow}$ \cite{Real2019}   &  26.0  & 8.5   & 5.3    & 555 \\
PNAS $^{\Uparrow}$ \cite{liu2018progressive}   &  26.0  & 8.5   & 5.3    & 555 \\
\midrule
\textbf{Micro-L} $^{\Uparrow}$                  & $\boldsymbol{27.6}$  & $\boldsymbol{9.4}$   & $\boldsymbol{3.7}$                  &
$\boldsymbol{363}$ \\\bottomrule               
\end{tabular}

\label{imagenet}
\end{table}

To validate the practicality of our output architectures, we experiment with the transferability of the architecture \textbf{Micro-L} from CIFAR-10 to ImageNet \cite{imagenet_cvpr09}. ImageNet dataset, which substantially shows its importance in real-world applications, contains more than one million images. With various resolutions, these images unevenly distribute in 1K categories. The general idea of transferring is to scale the architectures with a greater number of channels but a smaller number of layers, which is introduced by some classic works in NAS \cite{Zoph2018,Xie2019}. More specifically, the architectures starts with three stem convolutional layers with stride 2, which downsample the resolution by eight times. Following, there are 14 layers and 48 initial channels, where the reduction cells appear on the fifth and ninth layer. Some common data augmentation techniques are also adopted, including the random resize, the random crop, the random horizon flip, and the color jitter. We train our model with the SGD optimizer with 250 epochs, batch size of 1024, and resolution of $224 \times 224$ on 4 Nvidia Tesla V100 GPU. The initial learning rate is set to 0.5 and decays to $1 \times 10^{-5}$ linearly. In addition, the warmup strategy is applied on the first five epochs, increasing the learning rate from 0 to 0.5 linearly. The label smooth technique with a smooth rate of 0.1 is also adopted. Table. \ref{imagenet} shows the experimental results and the comparisons to state-of-the-art methods. It shows that our approach has superior performance comparing to the hand-crafted architectures and has competitive performance with state-of-the-art NAS algorithms.